\documentclass[letterpaper]{article}
\usepackage{proceed2e}
\usepackage[margin=1in]{geometry}

\usepackage{times}

\usepackage{microtype}
\usepackage{graphicx}
\usepackage{subfigure}
\usepackage{booktabs} % for professional tables

\usepackage{hyperref}

\usepackage[square,sort, round]{natbib}

\usepackage{algorithm} %algorithm environments
\usepackage{algorithmic}
\usepackage{color} % add color to text
\usepackage{amsthm} %for theorem environments
\usepackage{amsmath} %for equations
\usepackage{amsfonts} % for mathbb
\usepackage{tikz} % to draw markov chains
\usetikzlibrary{automata, positioning, shapes.geometric}
\usepackage{caption} %to add caption on the chain figure
\usepackage{mathtools} % to use multline for long equations
\usepackage{tabularx} % adapt tables to page margins

\usepackage{amssymb}% http://ctan.org/pkg/amssymb
\usepackage{pifont}% http://ctan.org/pkg/pifont
\newcommand{\cmark}{\ding{51}}%
\newcommand{\xmark}{\ding{55}}%

\makeatletter
\def\hlinewd#1{%
  \noalign{\ifnum0=`}\fi\hrule \@height #1 \futurelet
   \reserved@a\@xhline}
\makeatother

\newtheorem{proposition}{Proposition}[section] 
\newtheorem{theorem}{Theorem}[section] 
\newtheorem{corollary}{Corollary}[section] 
\newtheorem{assumption}{Assumption}[section] 
\newtheorem{lemma}{Lemma}[section] 
\newtheorem{definition}{Definition}[section] 

\title{Soft-Robust Actor-Critic Policy-Gradient}

\author{
  Esther Derman \\
  Technion, Israel\\
  \texttt{estherderman@technion.ac.il} 
  \And
  Daniel J. Mankowitz \\
  Technion, Israel\\
  \texttt{danielm@campus.technion.ac.il} 
  \AND
  Timothy A. Mann \\
  Google Deepmind, UK \\
  \texttt{timothymann@google.com} 
  \And
  Shie Mannor \\
  Technion, Israel\\
  \texttt{shie@ee.technion.ac.il} \\
}

\begin{document}
\maketitle

\begin{abstract}
Robust Reinforcement Learning aims to derive an optimal behavior that accounts for model uncertainty
in dynamical systems. However, previous studies have shown that by considering the worst case scenario, 
robust policies can be overly conservative. Our \textit{soft-robust} framework is an attempt to overcome
this issue. In this paper, we present a novel Soft-Robust Actor-Critic algorithm (SR-AC). 
It learns an optimal policy with respect to a distribution over an uncertainty set and stays robust to model uncertainty but avoids
the conservativeness of robust strategies. We show the convergence of SR-AC and test the efficiency of our approach
on different domains by comparing it against regular learning methods and their robust formulations.
\end{abstract}

\section{INTRODUCTION}
\label{intro}
Markov Decision Processes (MDPs) are commonly used to model sequential decision making in stochastic environments. 
A strategy that maximizes the accumulated expected reward is then considered as optimal and can be learned 
from sampling. However, besides the uncertainty that results from stochasticity of the environment,
model parameters are often estimated from noisy data or can change during testing \citep{biasMannor, modelmismatch}. 
This second type of uncertainty can significantly degrade the performance of the optimal strategy
from the model's prediction.  

Robust MDPs were proposed to address this problem \citep{iyengar, Nilim, scalingRMDP}.   
In this framework, a transition model is assumed to belong to a known uncertainty set and an optimal strategy
is learned under the worst parameter realizations. Although the robust approach is computationally efficient 
when the uncertainty set is state-wise independent, compact and convex, it can lead to overly conservative
results \citep{DRMDP1, DRMDP2, LDST, krectangularity}. 

%For example, consider an autonomous vehicle that must enter a traffic circle. If it is does not account for the different environments 
%he might be confronted to such as traffic and visibility, the aggressive agent
%may cause an accident under road conditions that are new to him. A robust agent would account for uncertainty
%by learning a strategy under the worst realization of all possible environments. However, when confronted to extreme scenarios
%such as high traffic under pouring rain, the robust agent may be stuck and not insert the road at all. Our claim is that one could
%relax this conservativeness and construct a softer behavior that keeps seeking robust actions. Ideally, the \textit{soft-robust} agent should
%stay agnostic to parameter uncertainty but still be able to insert the motorway properly.

For example, consider a business scenario where an agent's goal is to make 
as much money as possible. It can either create a startup which may make a fortune 
but may also result in bankruptcy. Alternatively, it can choose to live off school teaching and have almost no risk but low reward. 
%to work in an investment bank where it earns a lower income with significantly lower risk. 
%Third it could live off school teaching and have almost no risk but low reward. 
By choosing 
the teaching strategy, the agent may be overly conservative and not account for opportunities to 
invest in his own promising projects. Our claim is that one could
relax this conservativeness and construct a softer behavior that interpolates between being aggressive and robust. Ideally, the \textit{soft-robust} agent should
stay agnostic to outside financing uncertainty but still be able to take advantage of the startup experience. 

This type of dilemma can be found in various domains. In the financial market, investors seek 
a good trade-off between low risk and high returns regarding portfolio management \citep{finance}. 
In strategic management, product firms must choose the amount of resources they put into innovation. A conservative
strategy would then consist of innovating only under necessary conditions \citep{management}.

In this paper, we focus on learning a \textit{soft-robust policy} (defined below) by incorporating soft-robustness into 
an online actor-critic algorithm and show its convergence properties. Existing works mitigate conservativeness of robust MDP 
either by introducing coupled uncertainties \citep{LDST, krectangularity} or by assuming prior information on the uncertainty set \citep{DRMDP1, DRMDP2}.
They use dynamic programming techniques to estimate a robust policy.
However, these methods present some limiting 
restrictions such as non-scalability and offline estimation. Besides being computationally more efficient than batch learning \citep{RLOnlineOffline}, the use of an online algorithm is of significant interest in robust MDPs
because it can detect non-adversarial state-actions pairs along a trajectory and result in less conservative results, 
something which cannot be performed when solving the planning problem \citep{onlineRMDP}.
Other works have attempted to incorporate robustness into 
an online algorithm for policy optimization \citep{coherentRisk, robustOptions}. Although these approaches can deal with
large domains, a sampling procedure is required for each critic estimate in \citet{coherentRisk}, which differs from the strictly-speaking 
actor-critic. In \citet{robustOptions}, the authors introduce a robust version of actor-critic policy-gradient but
its convergence results are only shown for the actor updates.  Moreover, these works target the robust solution
which may be too conservative.
We review all existing methods in Section \ref{relatedWorks} and compare 
them to our approach.

To the best of our knowledge, our proposed work is the first attempt to incorporate a soft form of robustness into an online algorithm 
that has convergence guarantees besides being computationally scalable.
We deal with the curse of dimensionality by using function approximation that parameterizes 
the expected value within a space of much smaller dimension than the state space. 
By fixing a distribution over the uncertainty set, the induced soft-robust actor-critic 
learns a locally optimal policy in an online manner. Under mild assumptions on the set of distributions and uncertainty set,
we show that our novel Soft-Robust Actor-Critic (SR-AC) algorithm converges. 
We test the performance of soft-robustness on different domains, including a large state space with continuous actions. As far as we know, no other work has previously incorporated robustness into continuous action spaces. 

Our specific contributions are: (1) A soft-robust derivation of the objective function for policy-gradient; 
(2) An SR-AC algorithm that uses stochastic approximation to learn a variant of distributionally robust policy in an online manner;
(3) Convergence proofs of SR-AC;
(4) An experiment of our framework to different domains that shows the efficiency of soft-robust behaviors in a continuous action space as well. All proofs
can be found in the Appendix.

%%%%%%%%%%%%% NEW SECTION %%%%%%%%%%%%%%%%%%%
\section{BACKGROUND}
In this section, we introduce the background material related to our soft-robust approach.

\textbf{Robust MDP}
A robust MDP is a tuple $\langle \mathcal{X}, \mathcal{A}, r,  \mathcal{P}\rangle$ where $\mathcal{X}$ is a finite state-space, 
$\mathcal{A}$ is a finite set of actions, $r:\mathcal{X} \times \mathcal{A} \rightarrow \mathbb{R}$
is the immediate reward function which is deterministic and bounded and $\mathcal{P}$ is a set of transition matrices. We assume that $\mathcal{P}$ is structured as a cartesian product $\bigotimes_{x\in\mathcal{X}}\mathcal{P}_x$, 
which is known as the rectangularity assumption \citep{Nilim}. Given a state $x\in\mathcal{X}$, the uncertainty set $\mathcal{P}_x$ is a family of transition models $p_x\in\mathcal{P}_x$
we represent as vectors in which the transition probabilities of each action are arranged in the same block. 
For $x,y\in\mathcal{X}$ and $a\in\mathcal{A}$, denote by $p(x,a,y)$ the probability of getting from state $x$ to state $y$ given action $a$. 
%and $p(x,y)$ the probability of going from state $x$ to state $y$.

At timestep $t$, the agent is in state $x_t$ and chooses an action $a_t$ according to a stochastic policy $\pi: \mathcal{X} \rightarrow \mathcal{M}(\mathcal{A})$ that maps each state to 
a probability distribution over the action space, $ \mathcal{M}(\mathcal{A})$ denoting the set of distributions over $\mathcal{A}$. It then gets a reward $r_{t+1}$ and is brought to state $x_{t+1}$ with probability $p(x_t, a_t, x_{t+1})$. 

\textbf{Policy-Gradient}
Policy-gradient methods are commonly used to learn an agent policy. A policy $\pi$ is parametrized by $\theta$ and estimated by optimizing an objective function using stochastic gradient descent. A typical objective to be considered is the average reward function
\begin{equation*}
\begin{split}
J_p(\pi) &=  \lim_{T\rightarrow +\infty}\mathbb{E}^p[\frac{1}{T}\sum_{t = 0}^{T-1}r_{t+1}\mid \pi] \\
&= \sum_{x\in\mathcal{X}}d_p^{\pi}(x)\sum_{a\in \mathcal{A}}\pi(x,a)r(x,a)
\end{split}
\end{equation*}
where $r_t$ is the reward at time $t$, $p$ an aperiodic and irreducible transition model under which the agent operates and $d^\pi_p$ is the stationary distribution of the Markov process induced by $p$ under policy $\pi$. The gradient objective has previously been shown to be  
\[
\nabla_\theta J_p(\pi) = 
\sum_{x\in\mathcal{X}}d_p^\pi(x)\sum_{a\in\mathcal{A}}\nabla_\theta\pi(x,a)Q_p^\pi(x,a)
\]
where $Q_p^\pi(x,a)$ is the expected differential reward associated with state-action pair $(x,a)$. This gradient is then used to update the policy parameters according to: $\theta_{t+1} = \theta_t + \beta_t\nabla_\theta J_p(\pi)$, with $\beta_t$ a positive step-size \citep{PGFA}.

\textbf{Actor-Critic Algorithm} 
Theoretical analysis and empirical experiments have shown that regular policy-gradient methods present
a major issue namely high variance in the gradient estimates that results in slow convergence and inefficient sampling \citep{reviewAC}. 
First proposed by \citet{Barto}, actor-critic methods attempt to reduce the variance 
by using a critic that estimates the value function. 
They borrow elements from both value function and policy-based
methods. The value function estimate plays the role of a critic that helps evaluating the performance of the policy. As in
policy-based methods, the actor then uses this signal to update policy parameters in the direction of a gradient
estimate of a performance measure. Under appropriate conditions, the resulting algorithm is tractable and
converges to a locally optimal policy \citep{KondaTsikAC,ACBath}.

\textbf{Deep Q-networks}
Deep Q-Networks (DQNs) have proven their capability of solving complex learning tasks such as Atari video games \citep{atariDQN}. The Q-learning of \citet{Qlearning} typically learns
a greedy or $\epsilon$-greedy policy by updating the Q-function based on a TD-error. In Deep Q-learning \citep{atariDQN,dqnNature}, a non-linear function such as a neural network
is used as an approximator of the Q-function. It is referred to as a Q-network. The agent is then trained by optimizing the induced TD loss function thanks
to stochastic gradient descent. Like actor-critic, DQN is an online algorithm that aims at finding an optimal policy. The main difference with actor-critic is that it is \textit{off-policy}: 
it learns a greedy strategy while following an arbitrary behavior \citep{atariDQN}. 

\textbf{Deep Deterministic Policy-Gradient}
Since DQN acts greedily at each iteration, it can only handle small action spaces. The Deep Deterministic Policy-Gradient (DDPG) is an \textit{off-policy} algorithm that can learn behaviors in continuous action spaces \citep{DDPG}. It is based on an actor-critic architecture that follows the same baseline as in DQN. The critic estimates the current Q-value of the actor using a TD-error while the actor is updated according to the critic. This update is based on the chain rule principle which establishes equivalence between the stochastic and the deterministic policy gradient \citep{DPG}.

%%%%%%%%%%%%% NEW SECTION %%%%%%%%%%%%%%%%%%%
\section{SOFT-ROBUSTNESS}
\subsection{SOFT-ROBUST FRAMEWORK}
Unlike robust MDPs that maximize the worst-case performance, we fix a prior on how transition models are distributed over the uncertainty set. 
A distribution over $\mathcal{P}$ is denoted by 
$\omega$ and is structured as a cartesian product $\bigotimes_{x\in\mathcal{X}}\omega_x$. We find the 
same structure in \citet{DRMDP1, DRMDP2}.
Intuitively, $\omega$ can be thought as the way the adversary distributes over different transition models. 
The product structure then means that this adversarial distribution only depends on the current
state of the agent without taking into account its whole trajectory.
This defines a probability distribution $\omega_x$ over $\mathcal{P}_x$ independently for each state.

We further assume that $\omega$ is non-diffuse. 
This implies that the uncertainty set is non-trivial with respect to $\omega$
in a sense that the distribution does not affect zero mass to all of the models. 

\subsection{SOFT-ROBUST OBJECTIVE}
Throughout this paper, we make the following assumption:

\begin{assumption}
\label{a2_chain}
Under any policy $\pi$, the Markov chains resulting from any of the MDPs with transition laws $p \in \mathcal{P}$ are irreducible and aperiodic.
\end{assumption}

Define $d_p^\pi$ as the stationary distribution of the Markov chain that results from following policy
$\pi$ under transition model $p\in \mathcal{P}$. 

\begin{definition}
We call soft-robust objective or soft-robust average reward the function
$\bar{J}(\pi)  := \mathbb{E}_{p \sim \omega}  \left[   J_p(\pi)  \right] $.
%where 
%\begin{equation*}
%\begin{split}
%J_p(\pi) &=  \lim_{T\rightarrow +\infty}\mathbb{E}^p[\frac{1}{T}\sum_{t = 0}^{T-1}r_{t+1}\mid \pi] \\
%&= \sum_{x\in\mathcal{X}}d_p^{\pi}(x)\sum_{a\in \mathcal{A}}\pi(x,a)r(x,a),
%\end{split}
%\end{equation*}
%which is well defined according to Assumption \ref{a2_chain}.
\end{definition}

The distribution $\omega$ introduces a softer form of robustness in the objective function because it averages over
the uncertainty set instead of considering the worst-case scenario. It also 
gives flexibility over the level of robustness one would like to keep. A robust strategy would then consist of putting more mass on 
pessimistic transition models. Likewise, a distribution that puts all of its mass on one target 
model would lead to an aggressive behavior and result in model misspecification.

The \textit{soft-robust differential reward} is given by 
$\bar{Q}^\pi(x,a) := \mathbb{E}_{p\sim \omega} \left[Q_p^\pi(x,a)\right]$
where
\[Q_p^\pi(x,a) :=\mathbb{E}^{p}\biggl[\sum_{t = 0}^{+\infty}r_{t+1}-J_p(\pi)\vert x_{0}=x,a_0 = a, \pi\biggr].\]
Similarly, we introduce the quantity
\begin{equation*}
\begin{split}
\bar{V}^\pi(x)  :=\sum_{a\in \mathcal{A}}\pi(x,a) \bar{Q}^\pi(x,a)
= \mathbb{E}_{p\sim \omega} \left[V_p^\pi(x)\right]\\
\end{split}
\end{equation*}
with 
$V_p^\pi(x) := \sum_{a\in\mathcal{A}}\pi(x,a)Q_p^\pi(x,a).$
We will interchangeably term it as \textit{soft-robust expected differential reward} or \textit{soft-robust value function}.
%We finally introduce the \textit{soft-robust advantage function}: $\bar{A}^\pi(x,a):= \bar{Q}^\pi(x,a) - \bar{V}^\pi(x)$.

\subsection{SOFT-ROBUST STATIONARY DISTRIBUTION}
The above performance objective $\bar{J}(\pi)$ cannot as yet be written as an expectation of the reward over a stationary distribution because of
the added measure $\omega$ on transition models. Define the average transition model as $\bar{p} := \mathbb{E}_{p \sim \omega}[p]$.
It corresponds to the transition probability that results from distributing all transition models according to $\omega$. In
analogy to the transition probability that minimizes the reward for each given state and action in the robust transition function \citep{robustOptions}, our average 
model rather selects the expected distribution over all the uncertainty set for each state and action.
Under Assumption \ref{a2_chain}, we can show that the transition $\bar{p}$ as defined is irreducible and aperiodic, 
which ensures the existence of a unique stationary law we will denote by $\bar{d}^\pi$.

\begin{proposition}[Stationary distribution in the average transition model]
\label{statdistrib}
Under Assumption \ref{a2_chain}, the average transition matrix $\bar{p}:= \mathbb{E}_{p \sim \omega}[p]$ is irreducible and aperiodic. In particular, it admits a unique stationary distribution.
\end{proposition}

As in regular MDPs, the soft-robust average reward satisfies a Poisson equation, as
it was first stated in the discounted reward case in Lemma 3.1 of \citet{DRMDP1}. 
The following proposition reformulates this result for the average reward.

\begin{proposition}[Soft-Robust Poisson equation]
\label{softPoisson}
\begin{multline*}
\bar{J}(\pi) + \bar{V}^\pi(x)\\ 
= \sum_{a\in A}\pi(x,a) \left( r(x,a) + \sum_{x'\in \mathcal{X}} \bar{p}(x,a,x')\bar{V}^\pi(x')  \right)
\end{multline*}
\end{proposition}

This Poisson equation enables us to establish an equivalence between the expectation of the stationary distributions over the uncertainty set
and the stationary distribution of the average transition model, naming 
$\bar{d}^\pi (x)= \mathbb{E}_{p\sim \omega}[d^\pi_p(x)]$ with $x\in\mathcal{X}.$
Indeed, we have the following:

\begin{corollary}
\label{JbarDbar}
Recall $\bar{d}^\pi$ the stationary distribution for the average transition model $\bar{p}$. Then
\[\bar{J}(\pi) = \sum_{x\in\mathcal{X}}\bar{d}^\pi(x)\sum_{a\in\mathcal{A}}\pi(x,a)r(x,a).\] 
\end{corollary}

The goal is to learn a policy that maximizes the soft-robust average reward $\bar{J}$. We use
a policy-gradient method for that purpose.

%%%%%%%%%%%%% NEW SECTION %%%%%%%%%%%%%%%%%%%
\section{SOFT-ROBUST POLICY-GRADIENT}
In policy-gradient methods, we consider a class of parametrized stochastic policies
$\pi_\theta: \mathcal{X}\rightarrow \mathcal{M}(\mathcal{A})$ with $\theta \in\mathbb{R}^{d_1}$ 
and estimate the gradient of the objective function $\bar{J}$ with respect to policy parameters
in order to update the policy in the direction of the estimated gradient of $\bar{J}$. The optimal
set of parameters thus obtained is denoted by 
\[
\theta^*:= \arg\max_\theta \bar{J}(\pi_\theta).
\]
When clear in the context, we will omit the subscript $\theta$ in $\pi_\theta$ for notation ease. We further make the following assumption, which is standard in policy-gradient litterature:
\begin{assumption}
\label{a3_differentiable}
For any $(x,a)\in\mathcal{X}\times\mathcal{A}$, the mapping $\theta \mapsto \pi_\theta(x,a)$ is continuously differentiable with respect to $\theta$.
\end{assumption}

Using the same method as in \citet{PGFA}, we can derive the gradient of the soft-robust average reward thanks to the previous results. 

\begin{theorem}[Soft-Robust Policy-Gradient] 
\label{SRPGtheorem}
For any MDP satisfying previous assumptions, we have
\[
\nabla_{\theta}\bar{J}(\pi)=\sum_{x\in\mathcal{X}}\bar{d}^{\pi}(x)\sum_{a\in\mathcal{A}}\nabla_{\theta}\pi(x,a)\bar{Q}^{\pi}(x,a).
\]
\end{theorem}

In order to manage with large state spaces, we also introduce a linear approximation of $\bar{Q}^\pi$ we define as $f_w(x,a) := w^T\psi_{xa}$.
\citet{PGFA} showed that if the features $\psi_{xa}$ satisfy a compatibility condition and the approximation is locally optimal, then we can use it in place of $\bar{Q}^\pi$ and still point roughly 
in the direction of the true gradient. 

In the case of soft-robust average reward, this defines a soft-robust gradient update that possesses the ability to incorporate function approximation, as stated in the following result.
The main difference with that of \citet{PGFA} is that we combine the dynamics of the system with distributed transitions over the uncertainty set. 

\begin{theorem}[Soft-Robust Policy-Gradient with Function Approximation]
\label{SRPGFA}
Let $f_w: \mathcal{X}\times \mathcal{A} \rightarrow \mathbb{R}$ be a linear approximator of the soft-robust differential reward $\bar{Q}^\pi$.
If $f_w$ minimizes the mean squared error 
\begin{equation*}
\mathcal{E}^{\pi}(w):=\sum_{x\in\mathcal{X}}\bar{d}^{\pi}(x)\sum_{a\in \mathcal{A}}\pi(x,a)\biggl[\bar{Q}^{\pi}(x,a)-f_w(x,a)\biggr]^{2}
\end{equation*}

and is compatible in a sense that $\nabla _w f_w(x,a) = \nabla _\theta \log \pi(x,a)$, then

\begin{equation*}
\nabla _\theta\bar{J}(\pi) =\sum_{x\in \mathcal{X}}\bar{d}^{\pi}(x)\sum_{a \in \mathcal{A}}\nabla_\theta \pi(x,a)f_w(x,a)
\end{equation*}
\end{theorem}

We can further improve our gradient estimate by reducing its variance.
One direct method to do so is to subtract a baseline $b(x)$ from the previous gradient update. It is easy to show that this will not affect the gradient derivation. 
In particular, \citet{ACBath} proved that the value function minimizes the variance. It is therefore a proper baseline to choose. We can thus write the
following: 
\begin{equation}
\label{gradAdv}
\begin{split}
\nabla_\theta \bar{J}(\pi) &=\sum_{x\in \mathcal{X}}\bar{d}^{\pi}(x)\sum_{a \in \mathcal{A}} \nabla_\theta \pi(x,a)\biggl(\bar{Q}^\pi(x,a)  - \bar{V}^\pi(x) \biggr)\\
&=\sum_{x\in \mathcal{X}}\bar{d}^{\pi}(x)\sum_{a \in \mathcal{A}} \pi(x,a)\psi_{xa}\bar{A}^\pi(x,a),
\end{split}
\end{equation}
where $\bar{A}^\pi(x,a)$ is the \textit{soft-robust advantage function} defined by 
$\bar{A}^\pi(x,a):= \bar{Q}^\pi(x,a) - \bar{V}^\pi(x)$.

%%%%%%%%%%%%% NEW SECTION %%%%%%%%%%%%%%%%%%%
\section{SOFT-ROBUST ACTOR-CRITIC ALGORITHM}
%
%Policy parameters are then updated along the direction suggested
%by the critic, which corresponds to an approximate policy-gradient estimate. 

In this section, we present our 
SR-AC algorithm which is defined as Algorithm \ref{SRAC}. 
This novel approach incorporates a variation of distributional robustness into an online algorithm that effectively 
learns an optimal policy in a scalable manner. 
Under mild assumptions, the resulting two-timescale stochastic approximation algorithm converges to a locally optimal policy. 

\subsection{SR-AC ALGORITHM}
%As in regular actor-critic methods, policy parameters are updated along the direction of the soft-robust average reward gradient.
%In our setting, the soft-robust value function plays the role of the critic according to which the actor parameters are updated.
%We use linear temporal-difference to learn the critic online. 

An uncertainty set and a nominal model without uncertainty are provided as inputs. 
%These can be constructed from data sampling: take the nominal model as 
%the estimated parameters and construct the uncertainty set by random sampling inside the corresponding confidence interval. 
In practice, the nominal model and the uncertainty set can respectively be an estimate of the transition model resulting from
data sampling and its corresponding confidence interval. A distribution $\omega$ over the uncertainty set
is also provided. It corresponds to our prior information on the uncertainty set. 
The step-size sequences $(\alpha_t, \beta_t, \xi_t; t\geq 0)$  consist of small non-negative numbers properly chosen by the user (see Appendix for more details). 

At each iteration, samples are generated using the nominal model and the current policy.  
These are utilized to update the soft-robust average reward (Line 5) and the critic (Line 7) based on an estimate of 
a soft-robust TD-error we detail further. In our setting, the soft-robust value function plays the role of the critic according to which the actor parameters are updated. 
We then exploit the critic to improve our policy by updating the policy parameters
in the direction of a gradient estimate for the soft-robust objective (Line 8). This process is repeated until convergence.

\subsection{CONVERGENCE ANALYSIS}
We establish
convergence of SR-AC to a local maximum
of the soft-robust objective function
by following an ODE approach \citep{KushnerYin}. 

%We call \textit{soft-robust TD-error} at time $t$ the random quantity defined as 
%\[
%\delta_{t}:=r_{t+1}-\hat{J}_{t+1}+ \sum_{x' \in\mathcal{X}} \bar{p}(x_t, a_t, x')\hat{V}_{x'}-\hat{V}_{x_{t}}
%\]
%where $\hat{V}_{x_{t}}$ and $\hat{J}_{t}$ are unbiased estimates that satisfy $E[\hat{V}_{x_{t}}\mid x_t, \pi] = \bar{V}^\pi(x_t)$ and 
%$E[\hat{J}_{t+1}\mid x_t, \pi] = \bar{J}(\pi)$ respectively.
%We can easily show that this defines an unbiased estimate of the soft-robust advantage function $\bar{A}^\pi$ \cite{ACBath}.
%Thus, using equation (\ref{gradAdv}), an unbiased estimate of the gradient $\nabla_\theta \bar{J}(\pi)$ can be obtained by taking $\widehat{\nabla _\theta J}(\pi) = \delta_t\psi_{x_ta_t}$.

Consider $\hat{V}$ and $\hat{J}$ as unbiased estimates of $\bar{V}$ and $\bar{J}$ respectively. Calculating $\delta_t$ (Line 6 in Algorithm \ref{SRAC}) requires an estimate of the soft-robust average-reward that can be obtained by averaging over samples 
given immediate reward $r$ and distribution $\omega$ (Line 5).
In order to get an estimate of the soft-robust differential value $\hat{V}$, we use linear function
approximation. Considering $\varphi$ as a $d_2$-dimensional feature extractor over the state space $\mathcal{X}$, we may then
approximate $\bar{V}^\pi(x)$ as $v^T\varphi_x$, where $v$ is a $d_2$-dimensional parameter vector that we tune using linear TD.
This results in the following soft-robust TD-error:
\[
\delta_{t}:=r_{t+1}-\hat{J}_{t+1}+ \sum_{x' \in\mathcal{X}} \bar{p}(x_t, a_t, x')v_t^T\varphi_{x'}-v_t^T\varphi_{x_t},
\]
where $v_t$ corresponds to the current estimate of the soft-robust value function parameter.

As in regular MDPs, when doing linear TD learning, the function approximation of the value function introduces a 
bias in the gradient estimate \citep{ACBath}. Denoting it as $e^\pi$, we have 
$E[\widehat{\nabla _\theta J}(\pi)\mid \theta] = \nabla_\theta \bar{J} (\pi) + e^\pi$ (see Appendix).  
This bias term then needs to be small enough in order to ensure convergence.
 
Convergence of Algorithm \ref{SRAC} can be established
by applying Theorem 2 from \citet{ACBath} which exploits Borkar's work on two-timescale algorithms \citeyearpar{Borkar}. 
%For simplicity, we assume that the iterates resulting from the actor update (Line 8) in SR-AC remain bounded.
The convergence result is presented as Theorem \ref{convergence}.

\begin{theorem}
\label{convergence}
Under all the previous assumptions, given $\epsilon >0$, there exists $\delta>0$ such that for a parameter vector $\theta_t, t \geq 0$ obtained using the algorithm, 
if $\sup_{\pi_t}\lVert e^{\pi_t}\rVert<\delta$, then the SR-AC algorithm converges almost surely to an $\epsilon$-neighborhood of a local maximum of $\bar{J}$.
\end{theorem}

\begin{algorithm}[tb]
   \caption{SR-AC}
   \label{SRAC}
\begin{algorithmic}[1]
   \STATE {\bfseries Input:} $\mathcal{P}$ - An uncertainty set; 
$\hat{p}\in \mathcal{P}$ - A nominal model;
$\omega$ - A distribution over $\mathcal{P}$;
$f_x$ - A feature extractor for the SR value function; 

   \STATE{\bfseries Initialize:} $\theta = \theta_0$ - An arbitrary policy parameter; $v = v_0$ - An arbitrary set of value function parameters; $\alpha_0$, 
$\beta_0$, $\xi_0$ - Initial learning-rates; $x_0$ - Initial state\\

   \REPEAT
   \STATE  Act under $a_t \sim \pi_{\theta_t}(x_t, a_t) $\\
Observe next state $x_{t+1}$ and reward $r_{t+1}$ \\

    \STATE {\bfseries SR Average Reward Update: }  \\
    $ \hat{J}_{t+1} = (1-\xi_t)\hat{J}_{t} + \xi_tr_{t+1}$\\
    \STATE {\bfseries SR TD-Error: } \\
    $\delta_{t}=r_{t+1}-\hat{J}_{t+1}+ \sum_{x' \in\mathcal{X}} \bar{p}(x_t, a_t, x')\hat{V}_{x'}-\hat{V}_{x_{t}}$\\
    \STATE {\bfseries Critic Update: }  $v_{t+1} = v_t + \alpha_t\delta_t\varphi_{x_t}$\\
    \STATE {\bfseries Actor Update: } $\theta_{t+1} = \theta_t + \beta_t\delta_t\psi_{x_ta_t}$\\ 

   \UNTIL{convergence}
   
   \STATE {\bfseries Return: } SR policy parameters $\theta$ and SR value-function parameters $v$
\end{algorithmic}
\end{algorithm}

%%%%%%%%%%%%% NEW SECTION %%%%%%%%%%%%%%%%%%%
\section{NUMERICAL EXPERIMENTS}
\begin{figure}
\centering
\begin{tikzpicture}
   \node[state] at (3,0) (0) {$s_0$};
   \node[state] at (9,5) (1) {$F_1$};
   \node[state] at (9,3.5) (2) {$S_1$};
   \node[state] at  (9,2) (3) {$F_2$};
   \node[state] at (9,.5) (4) {$S_2$};
   \node[state] at (9,-1) (5) {$F_3$};
   \node[state] at (9,-2.5) (6) {$S_3$};
   
   \draw[every loop]
   	(0) edge[bend left] node[below, pos = .9, yshift = -1mm] {$a_1$, $R$ = -$10^5$} (1)
	(0) edge[bend left] node[below, pos = 0.8] {$a_1$, $R = 10^5$} (2)
	(0) edge[bend left] node[below, pos = 0.7] {$a_2$,$R = 0$ } (3)
	(0) edge[bend left] node[below, pos = 0.6] {$a_2$,$R = 2000$} (4)
	(0) edge[bend right] node[above, pos = 0.6,yshift = 1mm] {$a_3$,R = $-100$} (5)
	(0) edge[bend right] node[above, pos = 0.8] {$a_3$,$R = 5000$} (6);
\end{tikzpicture}
\caption{Illustration of the MDP with initial state $s_0$. States $F_1, F_2,F_3$ correspond
to failing scenarios for each action. The succeeding states are represented by states $S_1, S_2, S_3$.}
\label{bandit}
\end{figure}
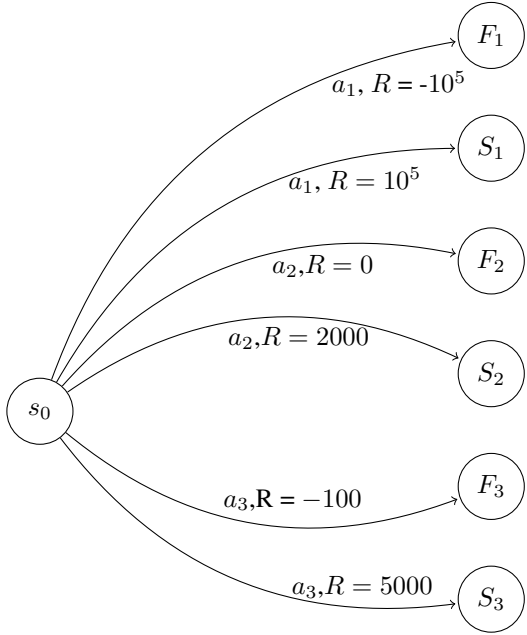

We demonstrate the performance of soft-robustness on various domains of finite as well as continuous state and action spaces. We used the existing structure of OpenAI Gym environments to run our experiments \citep{openai}.
 
\subsection{DOMAINS}
\textbf{Single-step MDP} 
We consider a simplified formulation of the startup vs teaching
dilemma described in Section \ref{intro}. The problem is modeled as a 
7-state MDP in which one action corresponds to one
strategy. An illustration of this construction is given in Figure \ref{bandit}. 
At the starting state $s_0$, the agent chooses one of three actions. Action $a_1$ [corresponds to the startup 
adventure] may lead it to a very high reward in case 
of success but can be catastrophic in case of failure. Action $a_2$ [corresponds to the teaching carrier] 
leads it to low positive reward in case of success with no possibility of negative reward. 
 Action $a_3$ [corresponds to an intermediate strategy]
%[corresponds to the investment banking carrier] 
can lead to an intermediate
 positive reward with a slight risk of negative reward.
Depending on the action it chose and if it succeeded or not, the agent is brought to 
one of the six right-hand states and receives the corresponding reward. It
is brought back to $s_0$ at the end of each episode.
We assume the probability of success to be the same for all three actions. 

\textbf{Cart-Pole} 
In the Cart-Pole system, the agent's goal consists of balancing a pole atop a cart in a vertical position.
It is modeled as a continuous MDP in which each state consists of a 4-tuple $\langle x, \dot{x},\theta, \dot{\theta} \rangle$
which represents the cart position, the cart speed, the pole angle with respect to the vertical and its angular speed respectively.   
The agent can make two possible actions: apply a constant force either to the right or to the left of the pole. It gets 
a positive reward of 1 if the pole has not fallen down and if it stayed in the boundary sides of the screen. If it
terminates, the agent receives a reward of 0. Since each episode lasts for 200 timesteps, the maximal
reward an agent can get is 200 over one episode.

\textbf{Pendulum} 
In the inverted pendulum problem, a pendulum starts in a random position and the goal is to swing it up so that it stabilizes upright. The state domain consists in a 2-tuple $\langle \theta, \dot{\theta} \rangle$ which represents the pendulum angle with respect to the vertical and its angular velocity. At each timestep, the agent's possible actions belong to a continuous interval $[-a, a]$ which represents the force level being applied. Since there is no specified termination, we establish a maximal number of $200$ steps for each episode. 

\subsection{UNCERTAINTY SETS}
For each experiment, we generate an uncertainty set $\mathcal{P}$ before training. In the single-step MDP, we sample from $5$ different probabilities of success using a uniform distribution
over $[0,1]$. In Cart-Pole, we sample 5 different lengths from a normal distribution
centered at the nominal length of the pole which we fix at $0.3$. We proceed similarly for Pendulum by generating 10 different masses of pendulum around a nominal mass of $2$. 
Each corresponding model thus generates a different transition function. We then sample
the average model by fixing $\omega$ as a realization of a Dirichlet distribution. A soft-robust update for the actor is applied 
by taking the optimal action according to this average transition function.

\subsection{LEARNING ALGORITHMS}
We trained the agent on the nominal model in each experiment. The soft-robust agent was learned using SR-AC in the single-step MDP. In Cart-Pole, we run a
soft-robust version of a DQN algorithm. The soft-robust agent in Pendulum was trained using a soft-robust DDPG.

\textbf{Soft-Robust AC}
We analyze the performance of SR-AC by training a soft-robust agent on the single-step MDP. We run a regular AC algorithm to derive an aggressive policy and learn a robust behavior by using a robust formulation of AC which consists in replacing the TD-error with a robust TD-error, as implemented in \cite{robustOptions}. The derived soft-robust agent is then compared with the resulting aggressive and robust strategies respectively.

\textbf{Soft-Robust DQN}
Robustness has already been incorporated in DQN \citep{kalman}.
The Q-network addressed there performs an online estimation of the Q-function by minimizing at each timestep $t$ the following robust TD-error:
\begin{equation*}
\begin{split}
\delta^{rob}_{dqn, t} :&=  r(x_t, a_t)- Q(x_t, a_t)  \\
&\quad+ \gamma \inf_{p\in \mathcal{P}} \sum_{x'\in\mathcal{X}} p(x_t, a_t, x')\max_{a'\in\mathcal{A}}  Q(x', a') ,
\end{split}
\end{equation*}
where $\gamma$ is a discount factor.

In our experiments, we incorporate a soft-robust TD-error inside a DQN that trains a soft-robust agent according to the induced loss function. The soft-robust TD-error for DQN is given by:
\begin{equation*}
\begin{split}
\delta^{srob}_{dqn, t} :&=   r(x_t, a_t) - Q(x_t, a_t)
\\
&\quad + \gamma \sum_{x'\in\mathcal{X}} \bar{p}(x_t, a_t, x')\max_{a'\in\mathcal{A}}  Q(x', a')
\end{split}
\end{equation*}
We use the Cart-Pole domain to compare the resulting policy with the aggressive and robust strategies that were obtained from a regular and a robust DQN respectively.

\textbf{Soft-Robust DDPG}
Define $\mu_t$ as the estimated deterministic policy at step $t$. We incorporate robustness in DDPG by updating the critic network according to the following robust TD-error:
\begin{equation*}
\begin{split}
\delta^{rob}_{ddpg,t} :&=  r(x_t, a_t)- Q(x_t, a_t)  \\
&\quad + \gamma \inf_{p\in \mathcal{P}} \sum_{x'\in\mathcal{X}} p(x_t, a_t, x')  Q(x', \mu(x_t)),
\end{split}
\end{equation*}
Similarly, we incorporate soft-robustness in DDPG by using the soft-robust TD-error:
\begin{equation*}
\begin{split}
\delta^{srob}_{ddpg,t} :&=   r(x_t, a_t) - Q(x_t, a_t)\\
&\quad + \gamma \sum_{x'\in\mathcal{X}} \bar{p}(x_t, a_t, x') Q(x', \mu_t(x_t))
\end{split}
\end{equation*}
We compare the resulting soft-robust DDPG with its regular and robust formulations in the Pendulum domain.

%In order to introduce robustness in that domain, we assume that the probability of success $\rho$
%is not precisely known. This induces an uncertainty set $\mathcal{P}(\rho)$ for transition models,
%where $\rho$ ranges between $0$ and $1$.  A probability law $\omega$ over the uncertainty set was
%then sampled from a Dirichlet distribution.

\subsection{IMPLEMENTATION}

For each experiment, we train the agent on the
nominal model but incorporate soft-robustness during learning. A soft-robust policy is learned thanks to SR-AC in the single-step MDP. We use a linear function approximation with 5 features to estimate the value function. For Cart-Pole, we run a DQN using a neural network of 3 fully-connected hidden layers 
with 128 weights per layer and ReLu activations. 
In Pendulum, a DDPG algorithm learns a policy based on two target networks: the actor and the critic network. Both have 2 fully-connected hidden layers with 400 and 300 units respectively. We use a $\tanh$ activation for the actor and a Relu activation for the critic output. 
We chose the ADAM optimizer to minimize all the induced loss functions. We used constant learning rates which worked well in practice. Each agent was trained over $3000$ episodes for the single-step MDP and Cartpole and tested over $600$ episodes per parameter setting. For Pendulum, the agents were trained over $5000$ episodes evaluated over $800$ episodes per parameter setting. Other hyper-parameter values can be found in the Appendix.

\subsection{RESULTS}
\begin{figure}[ht]
\vskip 0.2in
\begin{center}
\centerline{\includegraphics[width=\columnwidth]{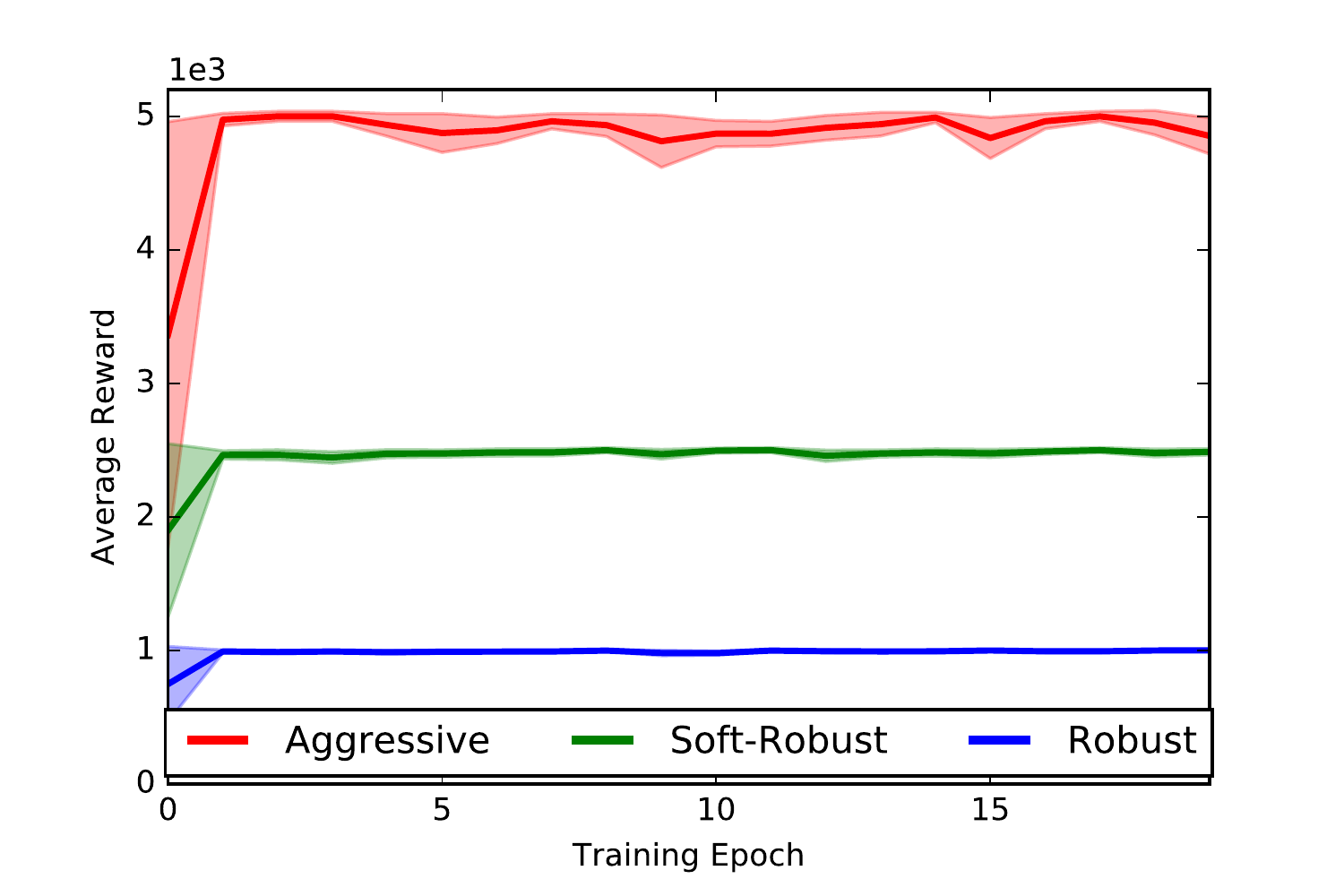}}
\caption{Comparison of robust, soft-robust and aggressive agents during training.
One training epoch corresponds to 300 episodes.}
\label{SRACtrain}
\end{center}
\vskip -0.2in
\end{figure}

\textbf{Single-step MDP} 
Figure \ref{SRACtrain} shows the evolution of the performance for all three agents during training. It becomes more stable along training time, which confirms convergence of SR-AC.
We see that the aggressive agent performs best due to the highest reward it can reach on the nominal model.
 The soft-robust agent 
%is less sensitive to failure and 
gets rewards in between the 
aggressive and the robust agent which performs the worst due to its pessimistic learning method. 

%Although this results in a more robust strategy, accounting for the overall uncertainty set without exclusively sticking to 
%the worst scenario leads to better performance than robust learning. Indeed, 
%in the case of soft-robustness, the weighting distribution puts mass on more optimistic transition models as well,
%which leads to higher rewards.

\begin{figure}[ht]
\vskip 0.2in
\begin{center}
\centerline{\includegraphics[width=\columnwidth]{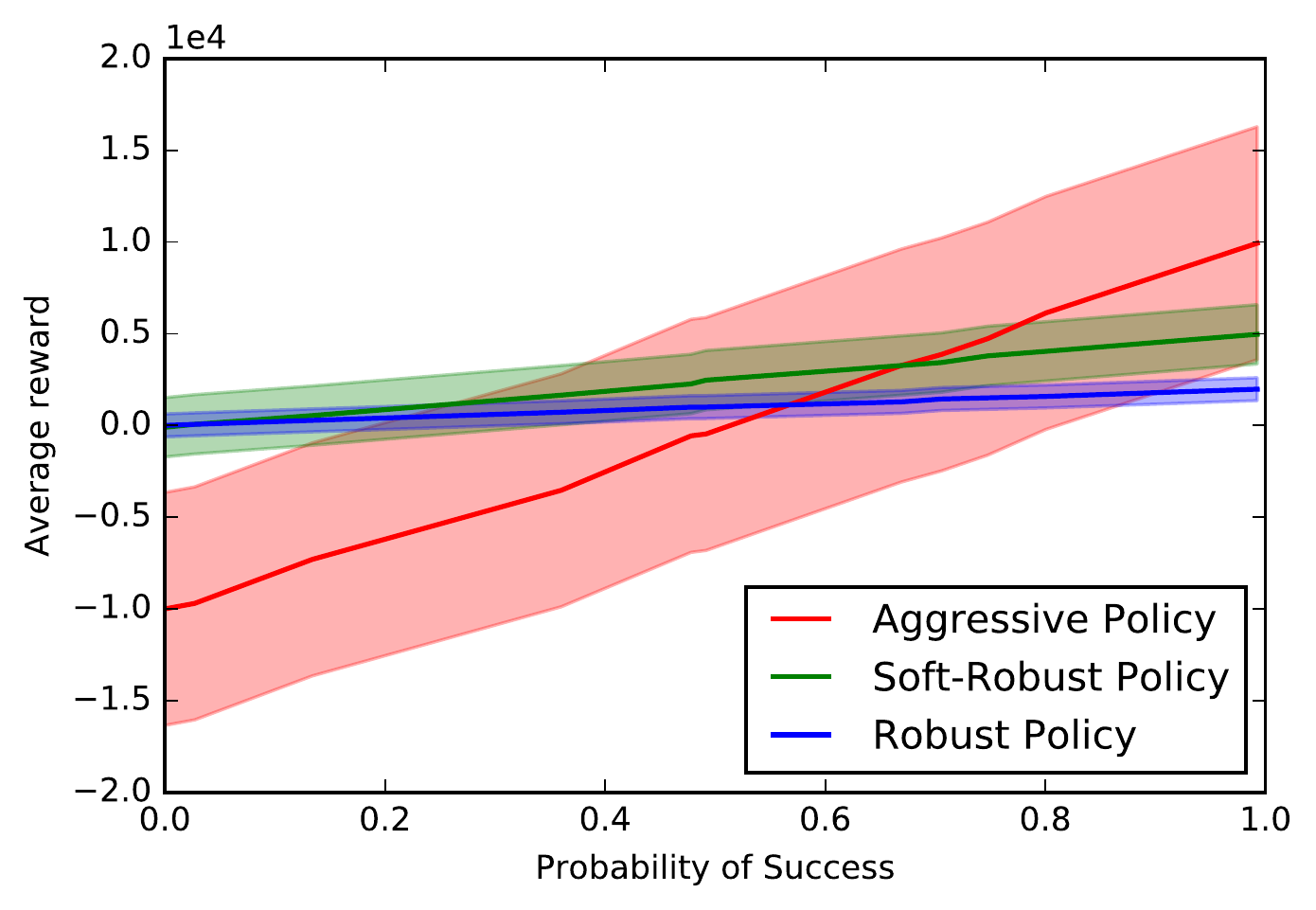}}
\caption{Average reward for AC, robust AC and SR-AC methods}
\label{SRACtest}
\end{center}
\vskip -0.2in
\end{figure}

The evaluation of each strategy is represented in Figure \ref{SRACtest}. As the probability 
of success gets low, the performance of the aggressive agent drops down below the robust and the soft-robust agents,
although it performs best
when the probability of success gets close to 1. The robust agent stays stable independently of the 
parameters but underperforms soft-robust agent which presents the best balance between high
reward and low risk. We noticed that depending on the weighting distribution initially set, soft-robustness tends to being more or less aggressive (see Appendix). Incorporating a distribution over the uncertainty set thus gives significant flexibility on the level of aggressiveness to be assigned to the soft-robust agent.

\begin{figure}[!!h]
\vskip 0.2in
\begin{center}
\centerline{\includegraphics[width=\columnwidth]{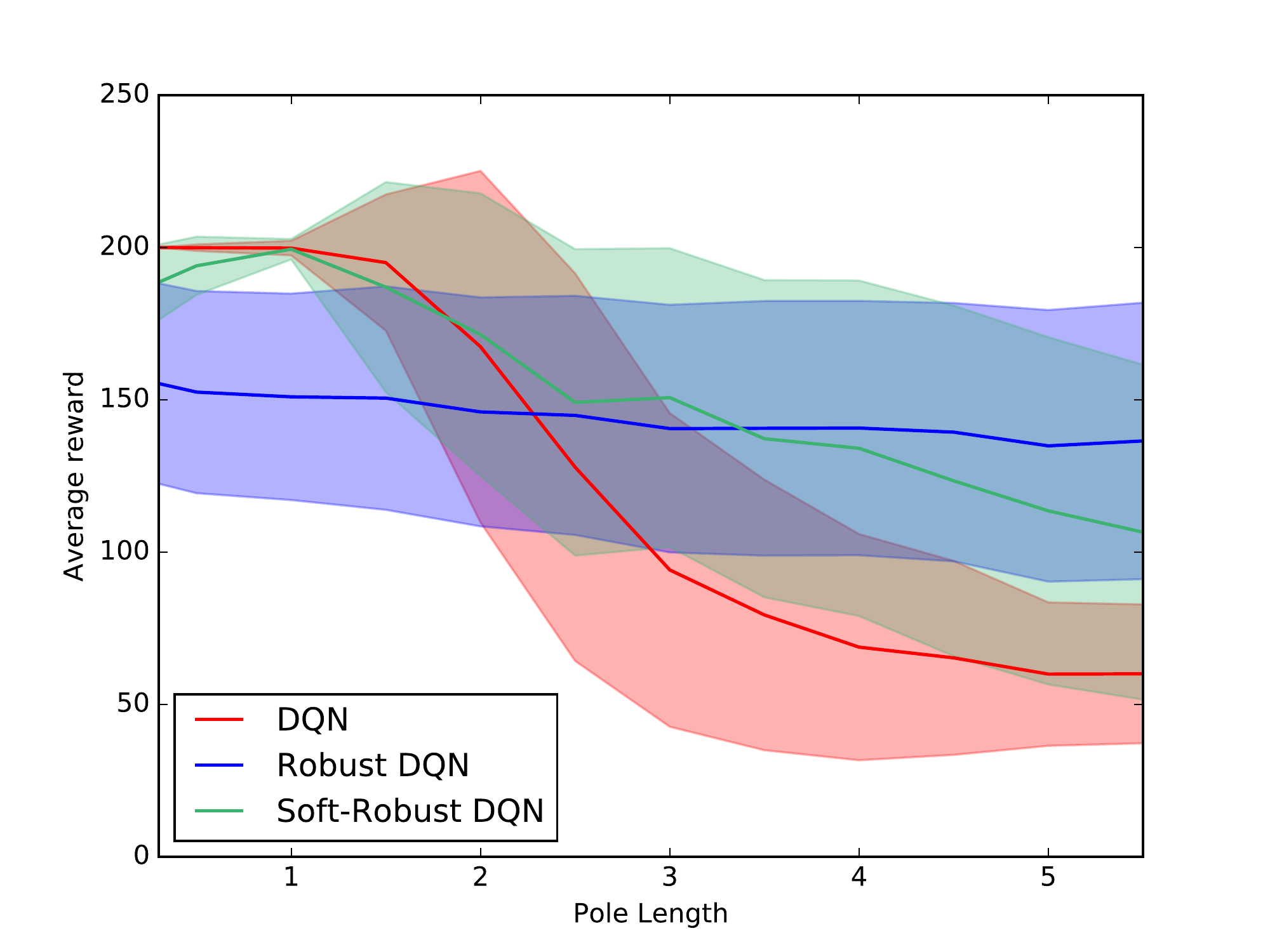}}
\caption{Average reward performance for DQN, robust DQN and soft-robust DQN}
\label{SRDQNtest1}
\end{center}
\vskip -0.2in
\end{figure}

\textbf{Cart-Pole} In Figure \ref{SRDQNtest1}, we show the performance of all three strategies over different values of pole length during testing. 
Similarly to our previous example, the non-robust agent performs well around the nominal model but its reward degrades on 
more extreme values of pole length. The robust agent keeps a stable reward under model uncertainty which is consistent with the results obtained in \cite{kalman, robustOptions}. 
However, it is outperformed by the soft-robust agent around the nominal model. Furthermore, the soft-robust strategy shows 
an equilibrium between aggressiveness and robustness thus leading to better performance than the non-robust agent on larger pole lengths. 
We trained a soft-robust agent on other weighting distributions and noted that depending on its structure, soft-robustness interpolates between aggressive and robust behaviors (see Appendix).

\begin{figure}[!!h]
\vskip 0.2in
\begin{center}
\centerline{\includegraphics[width=\columnwidth]{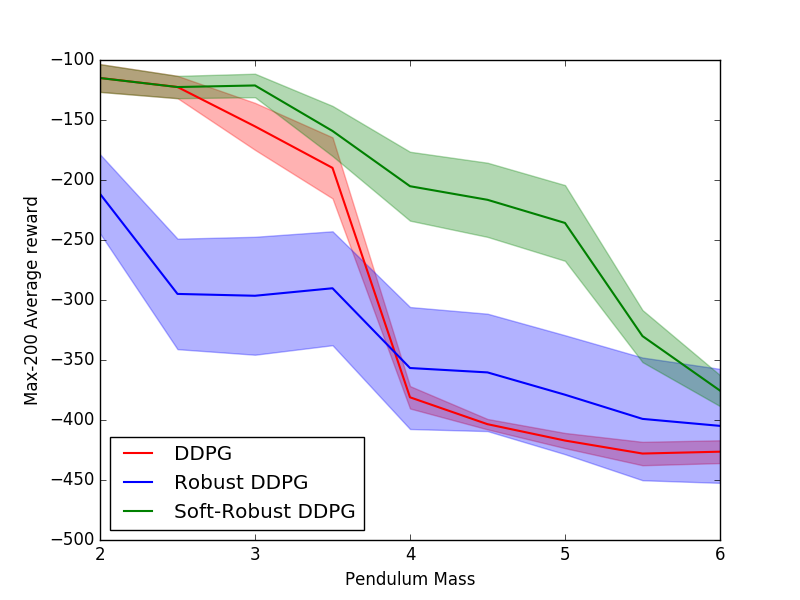}}
\caption{Max-200 episodes average performance for DDPG, robust DDPG and soft-robust DDPG}
\label{SRDDPG}
\end{center}
\vskip -0.2in
\end{figure}

\textbf{Pendulum}
Figure \ref{SRDDPG} shows the performance of all three agents when evaluating them on different masses. Since the performance among different episodes is highly variable, we considered the best $200$-episodes average reward as a performance measure. As seen in the figure, the robust strategy solves the task in a sub-optimal fashion, but is less affected by model misspecification due to its conservative strategy. The aggressive non-robust agent is more sensitive to model misspecification compared to the other methods as can be seen by its sudden dip in performance, below even that of the robust agent. The soft-robust solution strikes a nice balance between being less sensitive to model misspecification than the aggressive agent, and producing better performance compared to the robust solution.

\section{RELATED WORK}
\label{relatedWorks}

This paper is related to several domains in RL such as robust and distributionally robust MDPs,
actor-critic methods and online learning via stochastic approximation algorithms. 
Our work solves the problem of conservativeness encountered in robust MDPs by
incorporating a variational form of distributional robustness. The SR-AC algorithm combines scalability
to large scale state-spaces and online estimation of the optimal policy in an actor-critic algorithm.
Table \ref{table} compares our proposed algorithm with previous approaches.

%\begin{table}[!!h]
%\caption{Comparison of previous approaches with SR-AC}
%\begin{center}
%%{| l | l | l | l |}
%%{|l|l | l | l |p{0.8\linewidth}|}
%    \begin{tabularx}{\linewidth}{lX}
%    \hlinewd{2pt}
%     \textbf{Reference}          & \textbf{Scalable}  & \vtop{\hbox{\strut \textbf{Actor-}}\hbox{\strut \textbf{Critic}}}   &  \vtop{\hbox{\strut \textbf{Softly-}}\hbox{\strut \textbf{Robust}}}  \\ \hlinewd{2pt}
%    SR-AC (this paper)  & \cmark & \cmark & \cmark     \\ \hline
%     \cite{robustOptions}  & \cmark & \xmark & \xmark  \\ \hline
%     \cite{onlineRMDP} & \xmark&\xmark &  \xmark  \\ \hline
%     \cite{DRMDP2} & \xmark & \xmark & \cmark  \\ \hline
%    
%     \cite{krectangularity,LDST} & \xmark & \xmark & \xmark  \\ \hline 
%       \cite{coherentRisk} & \cmark & \xmark & \xmark   \\ \hline
%   
%   
%  
%   \cite{DRMDP1} & \xmark & \xmark & \cmark   \\ \hline
%    \cite{ACBath} & \cmark & \cmark & \xmark   \\\hline
%
%    \end{tabularx}
%\end{center}
%
%\label{table}
%\end{table}

\begin{table}[htbp]
\caption{Comparison of previous approaches with SR-AC}
\centering
%{| l | l | l | l |}
%{|l|l | l | l |p{0.8\linewidth}|}
    \begin{tabularx}{\linewidth}{|X|c|c|c|}
    \hlinewd{2pt}
   \textbf{Reference}        &  \textbf{Scalable}  & \vtop{\hbox{\strut \textbf{Actor-}}\hbox{\strut \textbf{Critic}}}  & \vtop{\hbox{\strut \textbf{Softly-}}\hbox{\strut \textbf{Robust}}}  \\  \hlinewd{2pt} 
    SR-AC (this paper)  & \cmark & \cmark & \cmark   \\ \hline
     \cite{robustOptions}  & \cmark & \xmark & \xmark \\   \hline
     \cite{onlineRMDP} & \xmark&\xmark &  \xmark \\ \hline
     \cite{DRMDP2} & \xmark & \xmark & \cmark \\  \hline
    
     \cite{krectangularity,LDST} & \xmark & \xmark & \xmark \\ \hline
       \cite{coherentRisk} & \cmark & \xmark & \xmark  \\ \hline
  
   \cite{DRMDP1} & \xmark & \xmark & \cmark    \\ \hline
    \cite{ACBath} & \cmark & \cmark & \xmark   \\ \hline
    \end{tabularx}
\label{table}
\end{table}

Many solutions have been addressed to mitigate conservativeness of robust MDP. 
\citet{LDST,krectangularity} relax the state-wise 
independence property of the uncertainty set and assume it to be coupled in a way such that the planning problem stays tracktable.
Another approach tends to assume
\textit{a priori} information on the parameter set. These methods include distributionally robust MDPs \citep{DRMDP1, DRMDP2}
in which the optimal strategy maximizes the expected reward under the most adversarial distribution
over the uncertainty set. For finite and known MDPs, under some structural assumptions on the considered set of distributions, 
this max-min problem reduces to classical robust MDPs and can be solved efficiently by dynamic programming \citep{putermanDP}. 

However,
besides becoming untracktable under large-sized MDPs, these methods use an offline learning approach which cannot
adapt its level of protection against model uncertainty and may lead to overly conservative results. 
The work of \citet{onlineRMDP} solutions this issue and addresses an online algorithm that learns the
transitions that are purely stochastic and those that are adversarial. Although it ensures less conservative results as well as low regret,
this method sticks to the robust objective while strongly relying on the finite structure of the state-space.
To alleviate the curse of dimensionality, we incorporate 
function approximation of the objective value and define it as a linear functional of features. 

First introduced in \citet{Barto} and later addressed by \citet{ACBath}, actor-critic algorithms are online learning methods that aim at finding an optimal policy. We used the formulation of \citet{ACBath} as a baseline for
the algorithm we proposed. The key difference between their work and ours is that we incorporate soft-robustness. This relates in a sense to the Bayesian Actor-Critic setup in which the critic returns a complete 
posterior distribution of value functions using Bayes' rule \citep{BAC, BRLSurvey, BayesianAC}. Our study keeps a frequentist approach, 
meaning that our algorithm updates return point estimates of the average value-function which prevents from tracktability issues besides enabling the distribution to be more flexible. Another major distinction is that the Bayesian approach incorporates a prior distribution on one model parameters whereas our method considers a prior on different transition models over an uncertainty set.

In \citet{coherentRisk, robustOptions}, the authors incorporate robustness into policy-gradient methods.
A sampling procedure is required for each critic estimate in \citet{coherentRisk}, which differs from the strictly-speaking 
actor-critic. A robust version of actor-critic policy-gradient is introduced in \citet{robustOptions} but its convergence guarantees are only shown
for robust policy-gradient ascent. Both of these methods target the robust strategy whereas we seek a soft-robust policy that is less conservative
while protecting itself against model uncertainty. 

%%%%%%%%%%%%% NEW SECTION %%%%%%%%%%%%%%%%%%%
\section{DISCUSSION}
We have presented the SR-AC framework that is able to learn policies which keep a balance between aggressive and robust behaviors. SR-AC requires a stationary distribution under the average 
transition model and compatibility conditions for deriving a soft-robust policy-gradient. 
We have shown that this ensures convergence of SR-AC. This is the first work that 
has attempted to incorporate a soft form of robustness into an online actor-critic method. Our approach has been shown to be computationally scalable to large domains because of its low computational price. In our experiments, 
we have also shown that the soft-robust agent interpolates between aggressive and robust strategies without being overly conservative
which leads it to outperform robust policies under model uncertainty even when the action space is continuous. Subsequent experiments should test the efficiency of soft-robustness on more complex domains. 

The chosen weighting
over the uncertainty set can be thought as the way the adversary distributes over different transition laws.
In our current setting, this adversarial distribution stays constant without accounting for the 
rewards obtained by the agent.
Future work should address the problem of learning the sequential game induced by 
an evolving adversarial distribution to derive an optimal soft-robust policy.
Other extensions of our work may also consider non-linear objective functions such as higher order moments with respect 
to the adversarial distribution.

\section*{Acknowledgements}
This work was partially funded by the Israel Science Foundation under contract 1380/16 and by the European Community's Seventh
Framework Programme (FP7/2007-2013) under grant agreement 306638 (SUPREL).

% In the unusual situation where you want a paper to appear in the
% references without citing it in the main text, use \nocite
% & ocite{langley00}

\bibliographystyle{plainnat}
\bibliography{SRACBib}

%\newpage
\onecolumn
\appendix
\section*{\hfil Appendix: Soft-Robust Actor-Critic Policy-Gradient  \hfil}
\section{Proofs}

\subsection{Proposition \ref{statdistrib}}
\begin{proof}
Fix $x,y\in \mathcal{X}$. For any policy $\pi$, we denote by $p(x,y)$ the probability of getting from state $x$ to state $y$, which can be written as $\mathbb{E}_{a\sim \pi(x)}[p(x,a,y)]$.
Since $\omega$ is non-diffuse, there exists $p_0$ such that $\omega(p_0) >0$. Also, by Assumption \ref{a2_chain}, there exists 
an integer $n$ such that $p_0^n(x,y)>0$. We thus have 
\begin{equation*}
\begin{split}
\bar{p}^n(x,y) &= \biggl(E_{p \sim \omega}[p]\biggr)^n(x,y)\\
\bar{p}^n(x,y)&\geq \biggl(p_0 \omega(p_0)\biggr)^n(x,y) \\
\bar{p}^n(x,y)&\geq p_0^n(x, y)\omega(p_0)^n >0
\end{split}
\end{equation*}
which shows that $\bar{p}$ is irreducible. Using the same reasoning, we show
$
\{n \in \mathbb{N} : p_0^n(x,x)>0\}   \subset \{n \in \mathbb{N} : \bar{p}^n(x,x)>0\} 
$
and then use the fact that $p_0$ is aperiodic to conclude that $\bar{p}$ is aperiodic too.
\end{proof}

%%%%%%%%%%%%%%%%%%%%%%%%%%%%%%%%%%%%%%%%%%%%%%%%%%%%%%%%%%%%%%%%%%%%%%%%%%%%%%%
%%%%%%%%%%%%%%%%%%%%%%%%%%%%%%%%%%%%%%%%%%%%%%%%%%%%%%%%%%%%%%%%%%%%%%%%%%%%%%%
\subsection{Proposition \ref{softPoisson}}
This recursive equation comes from the same reasoning as in Lemma 3.1 of \citet{DRMDP1}. We apply it to the average reward criterion.
\begin{proof}
For every $p\in\mathcal{P}$, we can apply the Poisson equation to the corresponding model:
\[
J_p(\pi) + V_p^\pi(x) = \sum_{a\in \mathcal{A}}\pi(x,a) \left( r(x,a) + \sum_{x'\in \mathcal{X}} p(x,a,x') V_p^\pi(x')  \right)
\]
By integrating with respect to $\omega$ we obtain:
\[
\bar{J}(\pi) + \bar{V}^\pi(x) 
= \sum_{a\in \mathcal{A}}\pi(x,a) \left( r(x,a) + \sum_{x'\in \mathcal{X}} E_{p \sim \omega}[p(x,a,x') V_p^\pi(x')]  \right)
\]
We then use the statewise independence assumption on $\omega$ to make the recursion explicit. We thus have
\begin{equation*}
\begin{split}
\bar{J}(\pi) + \bar{V}^\pi(x) 
&\overset{(1)}{=} \sum_{a\in A}\pi(x,a) \biggl( r(x,a) +\biggr. 
\biggl. \sum_{x'\in X} \int p(x,a,x') V_p^\pi(x') d\omega_x(p_x) d\omega_{x'}(p_{x'})  \biggr)\\
&\overset{(2)}{=} \sum_{a\in A}\pi(x,a) \left( r(x,a) + \sum_{x'\in X} E_{p_x \sim \omega_x}[p(x,a,x')] E_{p_{x'}\sim \omega_{x'}}[V_p^\pi(x') ] \right)\\
&= \sum_{a\in A}\pi(x,a) \left( r(x,a) + \sum_{x'\in X} \bar{p}(x,a,x') \bar{V}^\pi(x')  \right),
\end{split}
\end{equation*}
where $(1)$ results from the rectangularity assumption on $\omega$. 
$(2)$ Since $p(x,a,x')$ is an element of vector $p_x$ that only depends on the uncertainty set at state $x$ and $V_p^\pi(x')$ depends on the uncertainty set at state $x'$, we
can split the integrals. 
We slightly abuse notation here because a state can be visited multiple times. In fact, we implicitly introduce dummy states and treat
multiple visits to a state as visiting different states. More explicitely, we write $\omega$ as $\omega = \bigotimes_{t = 0}^{+\infty} \omega_{x,t}$ where $\omega_{x,t} = \omega_x$, $\omega_x$ being the distribution at state $x$. This representation is termed as the \textit{stationary model} in \citet{DRMDP1}.
\end{proof}

%%%%%%%%%%%%%%%%%%%%%%%%%%%%%%%%%%%%%%%%%%%%%%%%%%%%%%%%%%%%%%%%%%%%%%%%%%%%%%%
%%%%%%%%%%%%%%%%%%%%%%%%%%%%%%%%%%%%%%%%%%%%%%%%%%%%%%%%%%%%%%%%%%%%%%%%%%%%%%%
\subsection{Corollary \ref{JbarDbar}}
\begin{proof}
According to Proposition \ref{softPoisson} and summing up both sides of the equality with respect to the stationary distribution $\bar{d}^\pi$, we have
\begin{equation*}
\begin{split}
\bar{J}(\pi) + \sum_{x\in \mathcal{X}}\bar{d}^\pi(x)\bar{V}^\pi(x) 
&= \sum_{x\in \mathcal{X}}\bar{d}^\pi(x) \sum_{a\in \mathcal{A}}\pi(x,a) \left( r(x,a) + \sum_{x'\in \mathcal{X}} E_{p_x \sim \omega_x}[p(x,a,x')]\bar{V}^\pi(x')  \right)\\
&= \sum_{x\in \mathcal{X}}\bar{d}^\pi(x) \sum_{a\in \mathcal{A}}\pi(x,a) \left( r(x,a) + \sum_{x'\in \mathcal{X}} \bar{p}(x,a,x')\bar{V}^\pi(x')  \right)
\end{split}
\end{equation*}

Since $\bar{d}^\pi$ is stationary with respect to $\bar{p}$, we can then write
\[
\bar{J}(\pi) + \sum_{x\in \mathcal{X}}\bar{d}^\pi(x)\bar{V}^\pi(x)
 = \sum_{x\in \mathcal{X}}\bar{d}^\pi(x) \sum_{a\in \mathcal{A}}\pi(x,a) r(x,a) +  \sum_{x'\in \mathcal{X}} \bar{d}^\pi(x')\bar{V}^\pi(x'). 
\]

It remains to simplify both sides of the equality in order to get the result.
\end{proof}

%%%%%%%%%%%%%%%%%%%%%%%%%%%%%%%%%%%%%%%%%%%%%%%%%%%%%%%%%%%%%%%%%%%%%%%%%%%%%%%
%%%%%%%%%%%%%%%%%%%%%%%%%%%%%%%%%%%%%%%%%%%%%%%%%%%%%%%%%%%%%%%%%%%%%%%%%%%%%%%
\subsection{Theorem \ref{SRPGtheorem}}
We use the same technique as in \citet{PGFA,robustOptions} in order to prove a soft-robust version of policy-gradient theorem.
\begin{proof}
\begin{equation*}
\begin{split}
 \nabla_\theta\bar{V}^{\pi}(x) &= \nabla_\theta \sum_{a\in\mathcal{A}}\pi(x,a)\bar{Q}^{\pi}(x,a)\\
\nabla_\theta\bar{V}^{\pi}(x) &= \sum_{a\in\mathcal{A}}\biggl[ \nabla_\theta \pi(x,a)\bar{Q}^{\pi}(x,a)+\pi(x,a) \nabla_\theta\bar{Q}^{\pi}(x,a)\biggr]\\
\nabla_\theta\bar{V}^{\pi}(x)  &\overset{(1)}{=} \sum_{a\in\mathcal{A}}\biggl[ \nabla_\theta\pi(x,a)\bar{Q}^{\pi}(x,a)
+\pi(x,a) \nabla_\theta \biggl[r(x,a)-\bar{J}(\pi)+\sum_{x'\in\mathcal{X}}\bar{p}(x,a,x')\bar{V}^\pi(x')\biggr]\biggr]\\
\nabla_\theta\bar{V}^{\pi}(x)  &= \sum_{a\in\mathcal{A}}\biggl[\nabla_\theta \pi(x,a) \bar{Q}^{\pi}(x,a)+\pi(x,a)\biggl[-\nabla_\theta \bar{J}(\pi)+\sum_{x'\in\mathcal{X}}\bar{p}(x,a,x')\nabla_\theta \bar{V}^\pi(x') \biggr]\biggr]\\
\nabla_\theta \bar{J}(\pi)   &= \sum_{a\in\mathcal{A}}\biggl[\nabla_\theta \pi(x,a) \bar{Q}^{\pi}(x,a)+\pi(x,a)\biggl[\sum_{x'\in\mathcal{X}}\bar{p}(x,a,x')\nabla_\theta \bar{V}^\pi(x') \biggr]\biggr]-\nabla_\theta  \bar{V}^{\pi}(x)\\
\sum_{x\in\mathcal{X}} \bar{d}^{\pi}(x)\nabla_\theta \bar{J}(\pi)   &\overset{(2)}{=} \sum_{x\in\mathcal{X}} \bar{d}^{\pi}(x)\sum_{a\in\mathcal{A}}\biggl[\nabla_\theta \pi(x,a) \bar{Q}^{\pi}(x,a)+\sum_{a\in\mathcal{A}}\pi(x,a)\sum_{x'\in\mathcal{X}}\bar{p}(x,a,x')\nabla_\theta \bar{V}^\pi(x') \biggr]-\sum_{x\in\mathcal{X}} \bar{d}^{\pi}(x)\nabla_\theta \bar{V}^{\pi}(x)\\
\sum_{x\in\mathcal{X}} \bar{d}^{\pi}(x)\nabla_\theta \bar{J}(\pi)   &= \sum_{x\in\mathcal{X}} \bar{d}^{\pi}(x)\sum_{a\in\mathcal{A}} \nabla_\theta\pi(x,a) \bar{Q}^{\pi}(x,a)+\sum_{x\in\mathcal{X}} \bar{d}^{\pi}(x)\sum_{a\in\mathcal{A}}\pi(x,a)\sum_{x'\in\mathcal{X}}\bar{p}(x,a,x')\nabla_\theta \bar{V}^\pi(x') \\
   &\quad- \sum_{x\in\mathcal{X}} \bar{d}^{\pi}(x)\nabla_\theta \bar{V}^{\pi}(x)\\
\sum_{x\in\mathcal{X}} \bar{d}^{\pi}(x)\nabla_\theta  \bar{J}(\pi)   &\overset{(3)}{=} \sum_{x\in\mathcal{X}} \bar{d}^{\pi}(x)\sum_{a\in\mathcal{A}}\nabla_\theta \pi(x,a) \bar{Q}^{\pi}(x,a)+\sum_{x'\in\mathcal{X}} \bar{d}^{\pi}(x')\nabla_\theta \bar{V}(x')-\sum_{x\in\mathcal{X}} \bar{d}^{\pi}(x)\nabla_\theta \bar{V}^{\pi}(x)\\
\nabla_\theta \bar{J}(\pi)   &= \sum_{a\in\mathcal{A}}\nabla_\theta \pi(x,a)\bar{Q}^{\pi}(x,a)
\end{split}
\end{equation*}
\\
where $(1)$ occurs thanks to the soft-robust Poisson equation. $(2)$ Multiply both sides
of the Equation by $\sum_{x\in\mathcal{X}} \bar{d}^{\pi}(x)$. $(3)$ Since $ \bar{d}^{\pi}(x)$
is stationary with respect to $\bar{p}$, we have that $\sum_{x\in\mathcal{X}} \bar{d}^{\pi}(x)\sum_{a\in\mathcal{A}}\pi(x,a) \bar{p}(x,a,x')=\sum_{x'\in\mathcal{X}} \bar{d}^{\pi}(x')$.
\end{proof}

%%%%%%%%%%%%%%%%%%%%%%%%%%%%%%%%%%%%%%%%%%%%%%%%%%%%%%%%%%%%%%%%%%%%%%%%%%%%%%%
%%%%%%%%%%%%%%%%%%%%%%%%%%%%%%%%%%%%%%%%%%%%%%%%%%%%%%%%%%%%%%%%%%%%%%%%%%%%%%%
\subsection{Theorem \ref{SRPGFA}}

\begin{proof}
Recall the mean squared error:
\begin{equation*}
\mathcal{E}^{\pi}(w):=\sum_{x\in\mathcal{X}}\bar{d}^{\pi}(x)\sum_{a\in \mathcal{A}}\pi(x,a)\biggl[\bar{Q}^{\pi}(x,a)-f_w(x,a)\biggr]^{2}
\end{equation*}
with respect to the soft-robust state distribution $\bar{d}^{\pi}(x)$.
If we derive this distribution with respect to the parameters
$w$ and analyze it when the process has converged to a local optimum
as in \citet{PGFA}, then we get:
\[
\sum_{x\in\mathcal{X}}\bar{d}^{\pi}(x)\sum_{a\in \mathcal{A}}\pi(x,a)\biggl[\bar{Q}^{\pi}(x,a)-f_{w}(x,a)\biggr] \nabla_w f_{w}(x,a) =0
\]
\\
Additionally, the compatibility condition $\nabla _w f_w(x,a) = \nabla _\theta \log \pi(x,a)$ yields:
\begin{eqnarray*}
\sum_{x\in\mathcal{X}}\bar{d}^{\pi}(x)\sum_{a\in \mathcal{A}}\pi(x,a)\biggl[\bar{Q}^{\pi}(x,a)-f_{w}(x,a)\biggr]   \nabla _\theta \pi(x,a) \frac{1}{\pi(x,a)} & = & 0\\
\sum_{x\in\mathcal{X}}\bar{d}^{\pi}(x)\sum_{a\in \mathcal{A}} \nabla _\theta \pi(x,a) \biggl[\bar{Q}^{\pi}(x,a)-f_{w}(x,a)\biggr] & = & 0
\end{eqnarray*}
Subtract this quantity from the soft-robust policy gradient (Theorem \ref{SRPGtheorem}). We then have:
\begin{eqnarray*}
\nabla _\theta \bar{J}(\pi)& = & \sum_{x\in\mathcal{X}}\bar{d}^{\pi}(x)\sum_{a\in\mathcal{A}}\nabla _\theta \pi(x,a) \bar{Q}^{\pi}(x,a)-\sum_{x\in\mathcal{X}}\bar{d}^{\pi}(x)\sum_{a\in\mathcal{A}}\nabla _\theta \pi(x,a)\biggl[\bar{Q}^{\pi}(x,a)-f_{w}(x,a)\biggr]\\
 & = & \sum_{x\in\mathcal{X}}\bar{d}^{\pi}(x)\sum_{a\in\mathcal{A}}\nabla _\theta \pi(x,a) f_{w}(x,a).
\end{eqnarray*}
\end{proof}

%%%%%%%%%%%%%%%%%%%%%%%%%%%%%%%%%%%%%%%%%%%%%%%%%%%%%%%%%%%%%%%%%%%%%%%%%%%%%%%
%%%%%%%%%%%%%%%%%%%%%%%%%%%%%%%%%%%%%%%%%%%%%%%%%%%%%%%%%%%%%%%%%%%%%%%%%%%%%%%
\subsection{Convergence Proof for SR-AC}
We define as \textit{soft-robust TD-error} at time $t$ the following random quantity:
\[
\delta_{t}:=r_{t+1}-\hat{J}_{t+1}+ \sum_{x' \in\mathcal{X}} \bar{p}(x_t, a_t, x')\hat{V}_{x'}-\hat{V}_{x_{t}}
\]
where $\hat{V}_{x_{t}}$ and $\hat{J}_{t}$ are unbiased estimates that satisfy $E[\hat{V}_{x_{t}}\mid x_t, \pi] = \bar{V}^\pi(x_t)$ and 
$E[\hat{J}_{t+1}\mid x_t, \pi] = \bar{J}(\pi)$ respectively.
We can easily show that this defines an unbiased estimate of the soft-robust advantage function $\bar{A}^\pi$ \citep{ACBath}.
%\[\bar{A}^\pi(x,a):= \bar{Q}^\pi(x,a) - \bar{V}^\pi(x)\].
Thus, using equation (\ref{gradAdv}), an unbiased estimate of the gradient $\nabla_\theta \bar{J}(\pi)$ can be obtained by taking 
\[\widehat{\nabla _\theta J}(\pi) := \delta_t\psi_{x_ta_t}.\]

Similarly, recall the \textit{soft-robust TD-error} with linear function approximation at time $t$: 
\[
\delta_{t}:=r_{t+1}-\hat{J}_{t+1}+ \sum_{x' \in\mathcal{X}} \bar{p}(x_t, a_t, x')v_t^T\varphi_{x'}-v_t^T\varphi_{x_t},
\]
where $v_t$ corresponds to the current estimate of the soft-robust value function parameter.

As in regular MDPs, when doing linear TD learning, the function approximation of the value function introduces a bias in the gradient estimate \cite{ACBath}. 

Define the quantity 
\[
\widetilde{V}^{\pi}(x)   =  \sum_{a\in\mathcal{A}}\pi(x,a)\biggl[r(x,a)-\bar{J}(\pi)+\sum_{x'\in\mathcal{X}}\bar{p}(x,a,x')v_\pi^T\varphi_{x'}\biggr]
\]
where $v_\pi^T\varphi_{x'}$ is an estimate of the value function upon convergence of a TD recursion, that is $v_\pi = \lim_{t\rightarrow\infty}v_t$. Also, define as
$\delta_t^{\pi}$ the  associated error upon convergence:
 \[
 \delta_t^\pi := r_{t+1} - \hat{J}_{t+1}+ \sum_{x' \in\mathcal{X}} \bar{p}(x_t, a_t, x')v_\pi^T\varphi_{x'}-v_\pi^T\varphi_{x_t}. 
 \]

Similarly to Lemma 4 of \citet{ACBath}, the bias of the soft-robust gradient estimate is given by
 
 \[
 e^\pi := \sum_{x\in\mathcal{X}}\bar{d}^{\pi}(x)\biggl[\nabla_\theta\widetilde{V}^{\pi}(x)-\nabla_\theta v_\pi^T\varphi_{x}\biggr],
 \]
 
 that is $E[\widehat{\nabla _\theta J}(\pi)\mid \theta] = \nabla_\theta \bar{J} (\pi) + e^\pi$.
 This error term then needs to be small enough in order to ensure convergence of the algorithm.

Let denote as  $\bar{V}(v):=\Phi v$ the linear approximation to the soft-robust differential
value function defined earlier, where $\Phi\in\mathbb{R}^{n\times d_2}$
is a matrix and each feature vector $\varphi_{x}(k)$ corresponds to the $k^{th}$
column in $\Phi$. We make the following assumption:

\begin{assumption}
\label{a4_independance}
The basis functions $\varphi_{x}\in\mathbb{R}^{d_2}$
are linearly independent. In particular, $\Phi$ has full rank. We also have $\Phi v\neq e$ for all value function parameters 
$v\in\mathbb{R}^{d_2}$ where $e$ is 
a vector of all ones.
\end{assumption}

The learning rates $\alpha_{t}$ and $\beta_{t}$ (Lines 7 and 8 in Algorithm \ref{SRAC}) are established such that $\alpha_{t}\rightarrow0$
slower than $\beta_{t}\rightarrow0$ as $t\rightarrow\infty$. In
addition, $\sum_{t}\alpha_{t}=\sum_{t}\beta_{t}=\infty$ and $\sum_{t}\alpha_{t}^{2},\sum_{t}\beta_{t}^{2}<\infty$. 
We also set the soft-robust average reward step-size $\xi_t = c\alpha_t$ for a positive constant $c$.
The soft-robust average reward, TD-error and critic will all operate on the
faster timescale $\alpha_{t}$ and therefore converge faster. Eventually, define
a diagonal matrix $D$ where the steady-state distribution $\bar{d}^\pi$
forms the diagonal of this matrix. We write the soft-robust transition
probability matrix as:
\[
\bar{P}^\pi(x,x')=\sum_{a\in \mathcal{A}}\pi(x,a)\bar{p}(x,a,x'),
\]
where $x,x'\in \mathcal{X}$ and $\bar{p}$ designates the average transition model. By denoting $R^{\pi}\in\mathbb{R}^{n}$
the column vector of rewards $(\sum_{a\in\mathcal{A}} \pi(x_1, a)r(x_1, a), \dots \sum_{a\in\mathcal{A}} \pi(x_n, a)r(x_n, a))^T$ where
$(x_1,\dots, x_n)$ is a numbered representation of the state-space and using the following operator $T:\mathbb{R}^{n}\rightarrow\mathbb{R}^{n}$,
we can express the soft-robust Poisson equation as:
\[
T(J)=R^{\pi}-\bar{J}(\pi)e+\bar{P}^{\pi}J
\]
The soft-robust average reward iterates (Line 5) and the critic iterates (Line 7) defined in Algorithm \ref{SRAC} converge almost surely, as stated 
in the following Lemma which is a straightforward application of Lemma 5 from \citet{ACBath}.

\begin{lemma}
For any given policy $\pi$ and $\{\hat{J}_{t}\}, \{v_{t}\}$ as in the soft-robust average reward and critic updates, we have $\hat{J}_{t}\rightarrow \bar{J}(\pi)$ and $v_t\rightarrow v^\pi$ almost surely, where 
\begin{equation*}
\bar{J}(\pi)  = \sum_{x\in\mathcal{X}}\bar{d}^\pi(x)\sum_{a\in\mathcal{A}}\pi(x,a)r(x,a)
\end{equation*}
is the average reward under policy $\pi$ and $v^\pi$ is the unique solution to 
\[
\Phi^{T}D\Phi v^{\pi}=\Phi^{T}DT(\Phi v^{\pi})
\]
\end{lemma}

Thanks to all the previous results, convergence of Algorithm \ref{SRAC} can be established
by applying Theorem 2 from \citet{ACBath} which exploits Borkar's work on two-timescale algorithms \citeyearpar{Borkar}. 
For simplicity, we assume that the iterates resulting from the actor update (Line 10 of Algorithm \ref{SRAC}) in SR-AC remain bounded,
although one could prove convergence without such an assumption by incorporating an operator 
that projects any policy parameter to a compact set, as described in \citet{KushnerClark78}. 
The resulting actor update would then be the projected value of the predefined iterate. 
The convergence result is presented as Theorem \ref{convergenceA}.

\begin{theorem}
\label{convergenceA}
Under all the previous assumptions, given $\epsilon >0$, there exists $\delta>0$ such that for a parameter vector $\theta_t, t \geq 0$ obtained using the algorithm, 
if $\sup_{\pi_t}\lVert e^{\pi_t}\rVert<\delta$, then the SR-AC algorithm converges almost surely to an $\epsilon$-neighborhood of a local maximum of $\bar{J}$.
\end{theorem}

\section{Experiments}
\subsection{One-step MDP}

\begin{center}
    \begin{tabular}{| l | l | l | l |}
    \hlinewd{2pt}
    \textbf{Model Parameters} & \textbf{Value}  \\ \hlinewd{2pt}

    Nominal probability of success & $0.8$ \\ \hline
    Uncertainty set for probabilities of success & $[0.1, 0.7, 0.8, 0.3, 0.5]$ \\ \hline
    Weighting Distribution 1&  $[0.47, 0.22, 0.1, 0.09 , 0.12]$ \\ \hline
    Weighting Distribution 2&  $[0.63, 0.04, 0.05, 0.02, 0.26]$ \\ \hline
    Aggressive rewards &  $10^5 ; -10^5$ \\ \hline
    Soft robust rewards &  $5000 ; -100$ \\ \hline
    Robust rewards &  $2000 ; 0$ \\ \hline
    \end{tabular}
\end{center}

\begin{center} 
    \begin{tabular}{| l | l | l | l |}
    \hlinewd{2pt}
    \textbf{Hyperparameters} & \textbf{Value}  \\ \hlinewd{2pt}
    Critic Learning rate $\alpha$ & 5e-3 \\ \hline
    Actor Learning rate $\beta$ & 5e-5 \\ \hline
    Step size $\xi$ & $3\alpha$ \\ \hline
        Number of linear features & $5$ \\ \hline
    Number of episodes for training $M_{train}$& $3000$ \\ \hline
    Number of episodes for testing $M_{test}$& $600$ \\ \hline

    \end{tabular}
\end{center}

\begin{figure}[ht]
\vskip 0.2in
\begin{center}
\centerline{\includegraphics[width=.5\columnwidth]{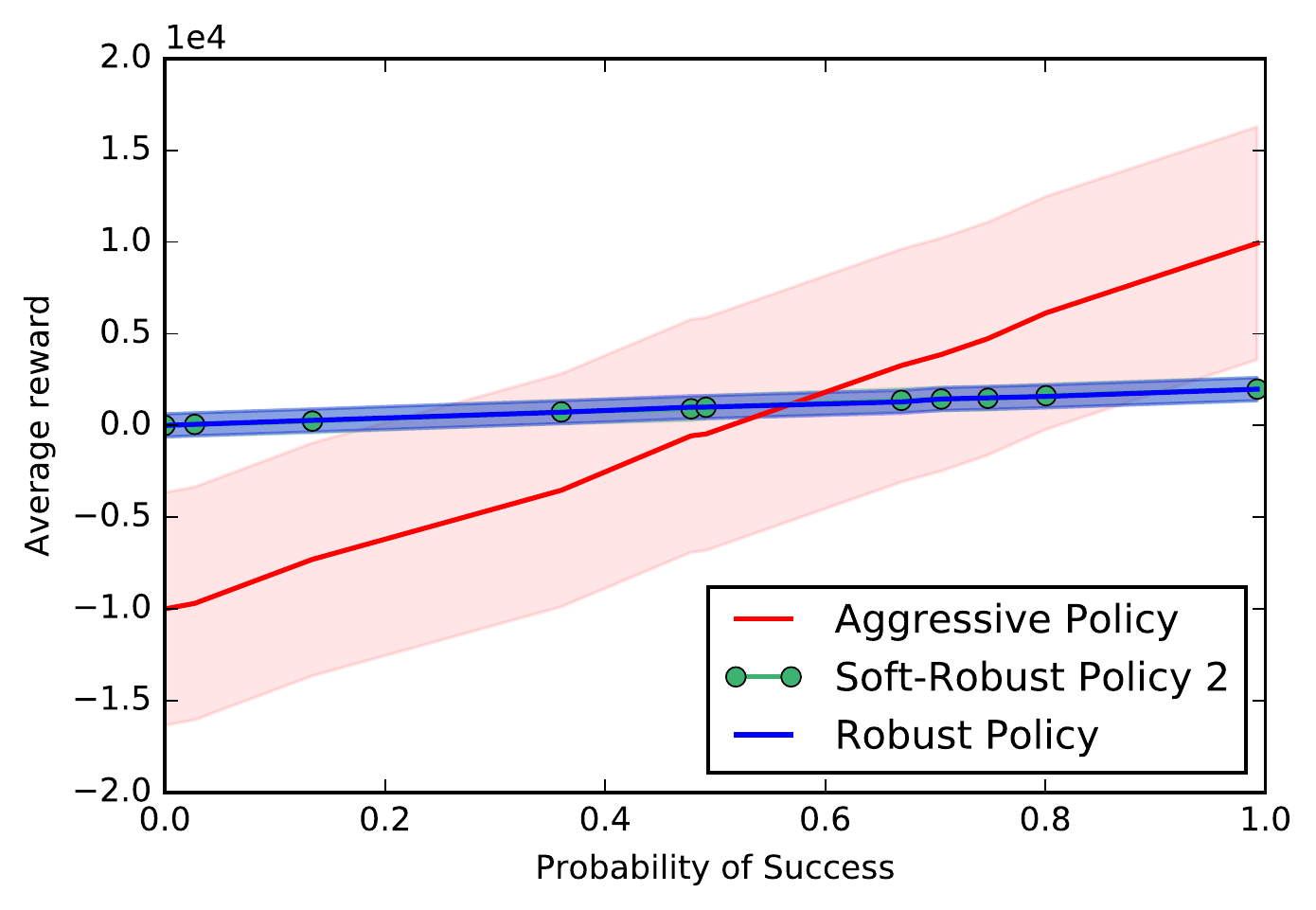}}
\caption{Average reward for different probabilities of success (distribution 2). Soft-robust
policy interpolates between aggressive and robust strategies.}
\label{SRACtest2}
\end{center}
\vskip -0.2in
\end{figure}

\subsection{Cart-Pole example}

\begin{center}
    \begin{tabular}{| l | l | l | l |} 
    \hlinewd{2pt}
    \textbf{Hyperparameters} & \textbf{Value}  \\ \hlinewd{2pt}
    Discount factor $\gamma$ & $0.9$ \\ \hline
    Learning rate & 1e-4 \\ \hline
    Mini-batch size & 256 \\ \hline
    Final epsilon & 1e-5\\ \hline
    Target update interval & $10$ \\ \hline
    Max number of episodes for training  $M_{train}$ & $3000$ \\ \hline
    Number of episodes for testing $M_{test}$ & $600$  \\ \hline

    \end{tabular}
\end{center}

\begin{figure}[ht]
\vskip 0.2in
\begin{center}
\centerline{\includegraphics[width=.5\columnwidth]{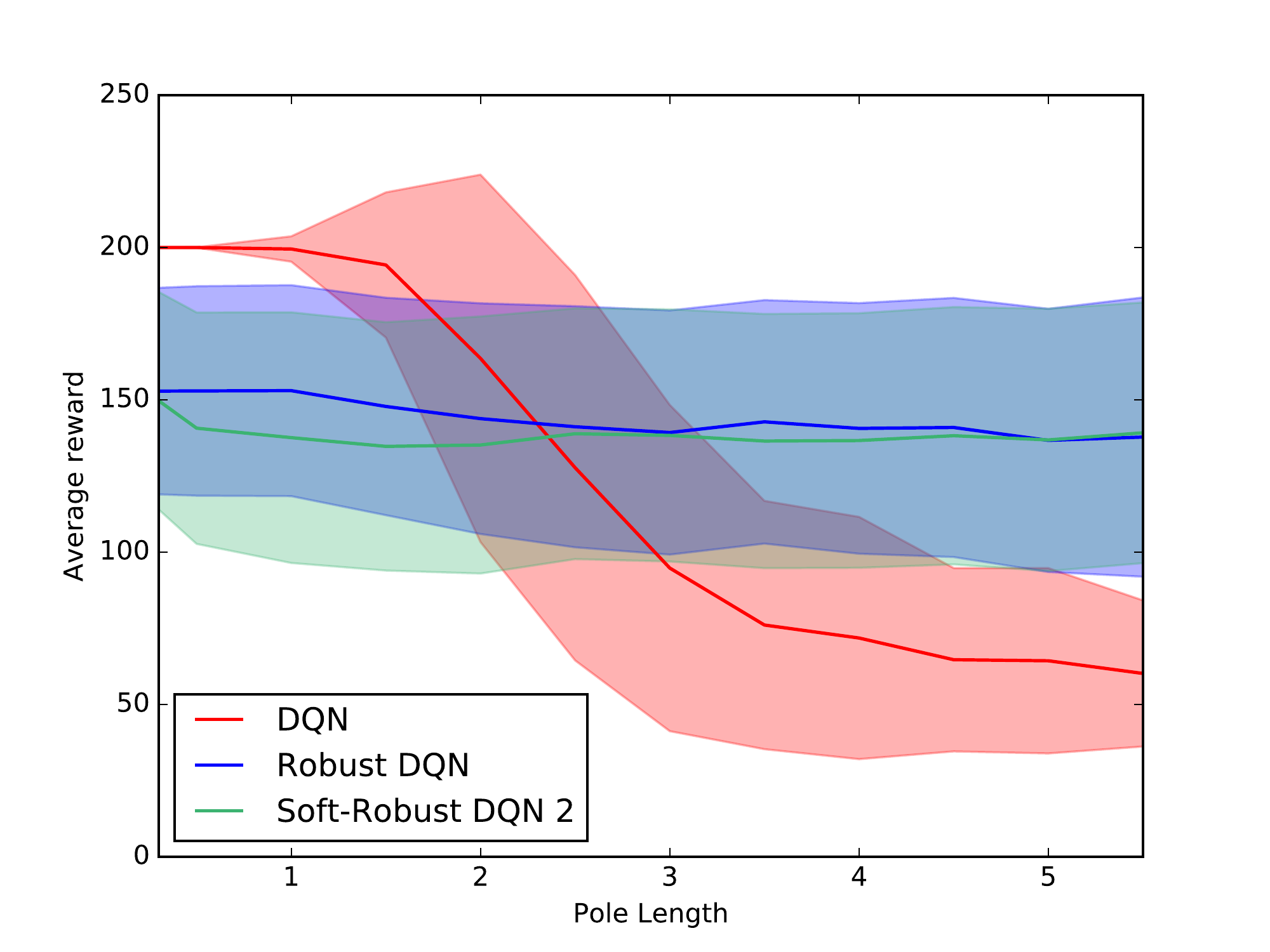}}
\caption{Average reward performance for DQN, robust DQN and soft-robust DQN (distribution 2). Soft-robust
policy interpolates between aggressive and robust strategies.}
\label{SRDQNtest2}
\end{center}
\vskip -0.2in
\end{figure}

We trained a soft-robust agent on a different weighting over the uncertainty set. Figure \ref{SRDQNtest2} shows the performance of 
the resulting strategy that presents a similar performance as the robust agent. This stronger form of robustness demonstrates the flexibility we have on the way 
we fix the weights, which leads to more or less aggressive behaviors.

\subsection{Pendulum}

\begin{center}
    \begin{tabular}{| l | l | l | l |}
    \hlinewd{2pt}
    \textbf{Hyperparameters} & \textbf{Value}  \\ \hlinewd{2pt}
    Discount factor $\gamma$ & $0.99$ \\ \hline
    Actor learning rate & 1e-5 \\ \hline
    Critic learning rate & 1e-3 \\ \hline
    Mini-batch size & 64 \\ \hline
    Soft target update & $\tau = 0.001$ \\ \hline
    Max number of episodes for training  $M_{train}$ & $5000$ \\ \hline
    Number of episodes for testing $M_{test}$ & $800$  \\ \hline

    \end{tabular}
\end{center}

% \begin{figure}[ht]
% \vskip 0.2in
% \begin{center}
% \centerline{\includegraphics[width=.5\columnwidth]{ddpg-v2.png}}
% \caption{Max-200 episodes average performance for DDPG, robust DDPG and soft-robust DDPG. The agents were trained with a different uncertainty set and a different distribution. In continuous spaces, the conservativeness}
% \end{center}
% \vskip -0.2in
% \end{figure}

%\bibliographystyle{plainnat}
%\bibliography{SRACBib}

\end{document}